\documentclass{article}
\pdfoutput=1

     \usepackage[preprint]{neurips_2022}

\usepackage[utf8]{inputenc} 
\usepackage[T1]{fontenc}    
\usepackage{hyperref}       
\usepackage{url}            
\usepackage{booktabs}       
\usepackage{amsfonts}       
\usepackage{nicefrac}       
\usepackage{microtype}      
\usepackage{xcolor}         
\usepackage{graphicx, subcaption, wrapfig}

\title{Transformers generalize differently\\from information stored in context vs weights}

\author{
    Stephanie C.Y. Chan\thanks{Equal contributions}\\DeepMind\\\And
    Ishita Dasgupta\footnotemark[1]\\DeepMind\\\And
    Junkyung Kim\\DeepMind\\\And
    Dharshan Kumaran\\DeepMind\\\And
    Andrew K. Lampinen\\DeepMind\\\And
    Felix Hill\\DeepMind}

\begin{document}

\maketitle

\begin{abstract}
Transformer models can use two fundamentally different kinds of information: information stored in weights during training, and information provided ``in-context'' at inference time. In this work, we show that transformers exhibit different inductive biases in how they represent and generalize from the information in these two sources. In particular, we characterize whether they generalize via parsimonious rules (\textit{rule-based generalization}) or via direct comparison with observed examples (\textit{exemplar-based generalization}). This is of important practical consequence, as it informs whether to encode information in weights or in context, depending on how we want models to use that information. In transformers trained on controlled stimuli, we find that generalization from weights is more rule-based whereas generalization from context is largely exemplar-based. In contrast, we find that in transformers pre-trained on natural language, in-context learning is significantly rule-based, with larger models showing more rule-basedness. We hypothesise that rule-based generalization from in-context information might be an emergent consequence of large-scale training on language, which has sparse rule-like structure. Using controlled stimuli, we verify that transformers pretrained on data containing sparse rule-like structure exhibit more rule-based generalization. 
\end{abstract}

\begin{wrapfigure}{R}{0.45\textwidth}
    \centering
    \begin{subfigure}[t]{0.25\textwidth}
        \caption{Partial exposure test.}
        \includegraphics[width=\textwidth]{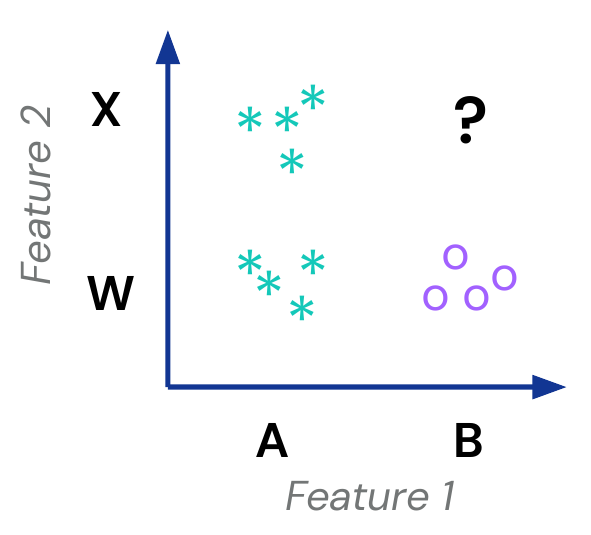}
        \label{fig:explainer:partial_exposure}
    \end{subfigure}
    
    \begin{subfigure}[t]{0.2\textwidth}
        \centering
        \caption{Rule-based.}
        \includegraphics[width=\textwidth]{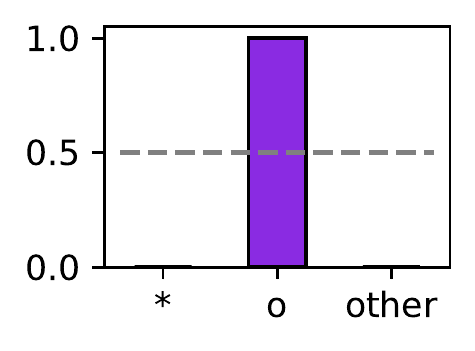}
        \label{fig:explainer:rule_based}
    \end{subfigure}
    \hfill
    \begin{subfigure}[t]{0.2\textwidth}
        \centering
        \caption{Exemplar-based.}  
        \includegraphics[width=\textwidth]{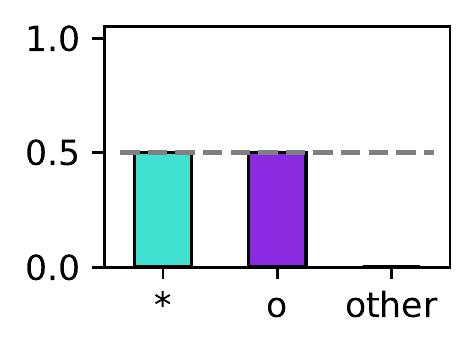}
        \label{fig:explainer:exemplar_based}
    \end{subfigure}
    \hfill
    \caption{
    Partial exposure test for differentiating rule-based vs exemplar-based generalization. Stimuli have two features. The model sees three combinations (\texttt{AX}, \texttt{AW}, and \texttt{BW}) in training or in context (depending on experiment), and is evaluated on a held-out (test) combination \texttt{BX}.
    \textbf{(\subref{fig:explainer:rule_based})} A rule-based model uses a parsimonious decision boundary that explains the data (here, based only on Feature 1), classifying the test as \texttt{o}. 
    \textbf{(\subref{fig:explainer:exemplar_based})} An exemplar-based model computes the similarity between test and training examples using all features. Since \texttt{BX} is equally similar to \texttt{AX} and \texttt{BW}, it is equally likely to classify it as \texttt{*} or \texttt{o}.
    }
    \label{fig:explainer}
    \vspace{-50pt}
\end{wrapfigure}

\section{Introduction}

Transformer-based architectures have an impressive ability to use both information stored in weights during training (``in-weights learning''), and information stored only in the inputs provided at inference time (without any gradient updates to the weights of the model; ``in-context learning'') \citep{chan_data_2022}. In-context learning on pretrained models 
enables learning efficiently from a few examples (``few-shot learning") \citep{brown_language_2020}, or even efficiently compressing a large dataset (``prompt tuning") \citep{li_prefix-tuning_2021, lester_power_2021, sun_black-box_2022}. Given the evident current and future potential for this new learning paradigm, it is important and useful to understand its inductive biases, especially how it differs from in-weights learning. 

One way to understand inductive bias is by examining how models \textit{generalize} to held-out data. In this work, we adapt the experimental paradigm in \cite{dasgupta_distinguishing_2022} that pose a classification task that distinguishes between two previously defined kinds of generalization behaviors (see \ref{fig:explainer}). A ``rule-based'' decision is made on the basis of minimal features that support the category boundary \citep{ashby_varieties_1986}, while an ``exemplar-based'' decision generalizes on the basis of similarity to examples from training data \citep{shepard_stimulus_1963}, invoking many or all features available. 

This distinction is particularly interesting when comparing in-weights vs in-context learning. Exemplar-based generalization (that uses all available features) is useful in a low-data regime where there is not enough information to form an abstract sparse rule \citep{feldman2020does}. On the other hand, sparser rule-based generalization may help avoid sensitivity to spurious correlation when training with large, noisy, naturalistic datasets (that are commonly used to train in-weights learning). 

We find that transformers exhibit a striking difference in their generalization from in-context vs in-weights information. 
Transformers display a strong inductive bias towards exemplar-based generalization from in-context information. In contrast, 
transformers display a strong inductive bias towards rule-based generalization from in-weights information.

However, when we pose a similar task to large transformer models pretrained on language, they exhibit stronger rule-based generalization from in-context information. 
One interpretation of these results is that the distribution of natural language is more compatible with rule-based generalization from context (rule-based generalization is in fact optimal in compositional domains like langauge \citep{arjovsky2019invariant}), and such patterns might present strong enough learning pressure to overcome -- and even reverse -- transformers' inherent bias towards exemplar-based generalization from context. 


\section{Experimental Design}

We adapted the ``partial exposure'' paradigm from \cite{dasgupta_distinguishing_2022} where each stimulus has two features; only one of the features predicts the label. We evaluate how the model generalizes to a held-out combination (using sparse rules or similarity to exemplars), see Fig \ref{fig:explainer}. 

First, we explored generalization in transformers trained on controlled synthetic data, where we can examine generalization from both in-weights and in-context information and directly compare them. Second, we repeat this experiment on pretrained language models and characterize their in-context generalization. Finally, we compare the patterns observed and invesigate factors that explain the differences observed.

\section{Results}

\subsection{Trained-from-scratch transformers}
\label{sec:from_scratch}

\begin{figure*}[tb]
    \centering
    \begin{subfigure}[t]{0.25\textwidth}
        \centering
        \caption{From in-weights.}
        \vspace{10pt}
        \includegraphics[width=\textwidth]{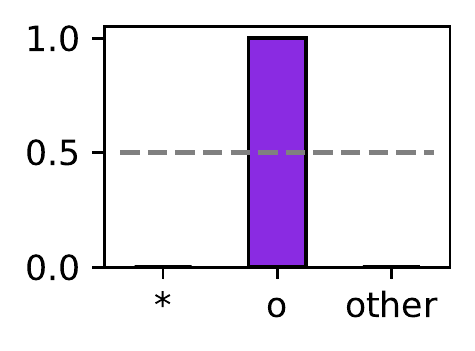}
        \label{fig:from_scratch:in-weights}
    \end{subfigure}
    \hfill
    \begin{subfigure}[t]{0.25\textwidth}
        \centering
        \caption{From in-context; few-shot training.}  
        \includegraphics[width=\textwidth]{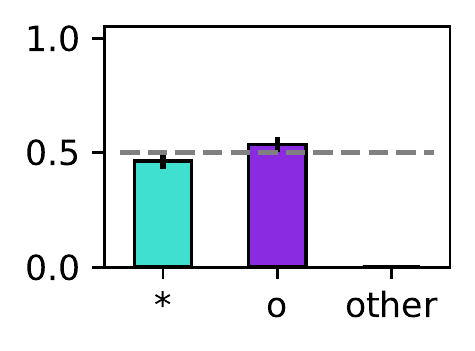}
        \label{fig:from_scratch:in-context}
    \end{subfigure}
    \hfill
    \begin{subfigure}[t]{0.25\textwidth}
        \centering
        \caption{From in-context; rule-based training.}  
        \includegraphics[width=\textwidth]{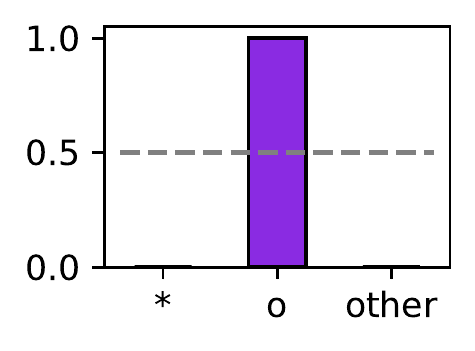}
        \label{fig:from_scratch:train_on_rule_based}
    \end{subfigure}
    \hfill
    \vskip-1em
 
    \caption{Generalization patterns of transformer models trained on synthetic data: frequency of various model outputs when presented with the held-out stimulus of the partial exposure paradigm (Fig \ref{fig:explainer}). \textbf{(\subref{fig:from_scratch:in-weights})} Generalization from weights is completely rule-based. 
    \textbf{(\subref{fig:from_scratch:in-context})} In contrast, generalization from context is exemplar-based. 
    \textbf{(\subref{fig:from_scratch:train_on_rule_based})} The exemplar-based bias in in-context learning can be overcome by pretraining the model on sequences that explicitly require rule-based generalization.}
    \label{fig:from_scratch}
    \vskip-0.5em
\end{figure*}

For the trained-from-scratch transformers, we passed sequences of stimulus-label pairs as inputs to the transformer model \citep{vaswani_attention_2017}. The sequences consisted of two parts: a context (24 tokens; i.e. 12 stimulus-label pairs) and a query (stimulus). The model was trained to minimize a softmax cross-entropy loss on the prediction for the final (query) stimulus. Each stimulus consists of two subvectors concatenated together into a single token (Fig \ref{app:fig:stimulus_examples}) -- these subvectors comprise the two features of the partial exposure paradigm. See Appendix \ref{app:from_scratch_details} for further details.

\paragraph{Generalization from in-weights information is rule-based.} To investigate generalization from in-weights information, we trained the model on partial exposure data, and evaluated on the held-out combination. During training, the label for each stimulus class was fixed, so that the stimulus-label mappings were stored in weights. The context tokens are uninformative for the query. After training, we measured the model's biases by evaluating on the held-out class combination. See Appendix Fig \ref{app:fig:in_weights_seqs} for details. When trained and evaluated in this way, transformers displayed fully rule-based generalization 
(Fig \ref{fig:from_scratch:in-weights}), i.e. based on a sparse rule that only took the first feature 
into account.

\paragraph{Generalization from in-context information is exemplar-based.} To investigate generalization from in-context information, we first pretrained the model for few-shot in-context learning (see \ref{app:pretrain_fsl} for more details), i.e. to refer to information provided in-context when making a query prediction. Importantly, pretraining for few-shot learning imparts no bias towards either rule-based or exemplar-based generalization, because either is an equally valid strategy for solving the few-shot problems. Then we evaluated the model on partial exposure sequences. See Appendix Fig \ref{app:fig:in_context_seqs} for example sequences used for training and evaluation. When trained and evaluated in this way, transformers displayed totally exemplar-based generalization, in striking contrast to the rule-based generalization that was observed from weights. That is, when queried on the held-out combination \texttt{BX}, the models were equally likely to output the labels associated with either \texttt{AX} and \texttt{BW} (Fig \ref{fig:from_scratch:in-context}), indicating that they are comparing the query directly (and along all features) to examples that were shown in context.


\subsection{Pretrained language models}
\label{sec:LLM}

Do these patterns hold when evaluating on pretrained language models, arguably the most-used transformer-based model at the moment? We used a large (70B parameters) pre-trained language model (LM) trained for autoregressive language prediction on large-scale web data \citep{Hoffmann2022TrainingCL}. To investigate in-context generalization in this model, we use the same ``partial exposure'' experimental paradigm, but instead using shape and color words as the two features comprising the stimuli, and nonsense words for the labels (see Fig \ref{app:fig:LM_seqs} for an example). We take the completion generated by the LM as the predicted query label.

With the synthetic data, we could ensure no a priori bias towards any feature; we have no such guarantee here. We account for any possible bias toward generalizing along a specific feature with a control condition from \citep{dasgupta_distinguishing_2022} where neither feature is more predictive in the training data, 
so the model must use pure exemplar-based generalization. Performance in this condition is now used as the baseline for pure exemplar-based generalization, and deviation from this baseline is the measure for rule-based generalization. See Appendix \ref{app:LLM_details} for further details. We don't investigate in-weights generalization in these models, since that would require manipulation and control of the training data and large-scale re-training.

\begin{figure*}[tb]
    \centering
    \begin{subfigure}[t]{0.25\textwidth}
        \centering
        \caption{Baseline feature bias.}  
        \includegraphics[width=\textwidth]{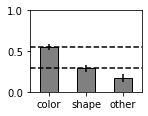}
        \label{fig:LM:control}
    \end{subfigure}
    \hfill
    \begin{subfigure}[t]{0.24\textwidth}
        \centering
        \caption{Shape predictive.}  
        \includegraphics[width=\textwidth]{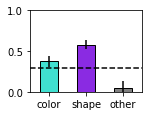}
        \label{fig:LM:shape}
    \end{subfigure}
    \hfill
    \begin{subfigure}[t]{0.24\textwidth}
        \centering
        \caption{Color predictive.}  
        \includegraphics[width=\textwidth]{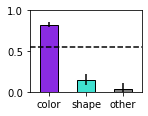}
        \label{fig:LM:color}
    \end{subfigure}
    \hfill
    \begin{subfigure}[t]{0.24\textwidth}
        \centering
        \caption{Effect of model size.}  
        \includegraphics[width=\textwidth]{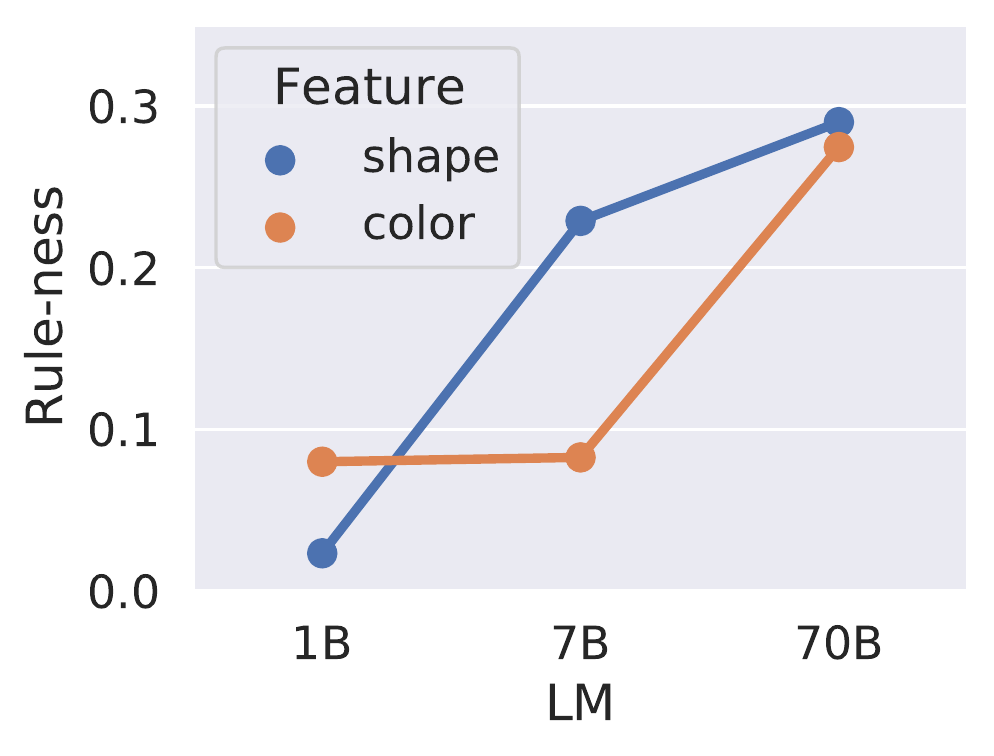}
        \label{fig:LM:sizes}
    \end{subfigure}
    \hfill
    \vskip-1em
    \caption{Generalization from in-context information in a pretrained LM. We classify LM responses by whether it gives the label consistent with generalizing along color, shape, or neither. 
    \textbf{(\subref{fig:LM:control})} Measuring feature-level bias with the Control condition; the model prefers to generalize along color. We use these results as baselines for the partial exposure conditions. \textbf{(\subref{fig:LM:shape})} When a sparse rule-based decision boundary supports shape as predictive, the model classifies along shape more often than in the baseline control (dotted line). \textbf{(\subref{fig:LM:shape})} Similarly when color is predictive, the model classifies along color more often than in the baseline control (dotted line). \textbf{(\subref{fig:LM:sizes})} Smaller LMs are less rule-based.}
    \label{fig:LM}
    \vskip-1em
\end{figure*}

\paragraph{In language models, generalization from in-context information is partially rule-based.} First, in the control condition where the model is forced to use exemplar-based generalization, we find that the language model prefers generalizing along the color dimension rather than the shape dimension (Fig \ref{fig:LM:control}). Since bias towards shape gives better classification performance on naturalistic visual stimuli \citep{landau1988importance, geirhos2018imagenet}, it is interesting that we see the opposite bias in a system trained on naturalistic text data; future work should look into possible explanations (e.g. the ``pragmatic'' nature of language; \cite{degen_when_2019}). The LM predominantly produces one of the two labels provided in context, but does sometimes produce an unseen word (`other' in Fig \ref{fig:LM}). The LM produces an unseen word more frequently when the query is also unseen in-context (not reported), suggesting a mutual exclusivity bias \citep{gandhi2020mutual}, also worth future investigation.

To measure the degree of rule-based generalization, we compared each partial exposure result to the respective control: for instance, we compared the probability of generalizing along color in the color predictive partial exposure evaluation condition (Fig \ref{fig:LM:color}) vs in the control condition. If a model uses pure exemplar-based generalization, there will be no difference in the classification pattern between the partial exposure and the control conditions. Alternatively, rule-based generalization predicts an increased sensitivity to the predictive feature dimension (either shape or color) compared to control. 
Here, we found evidence of rule-based generalization for both color and shape. This is still in contrast with the purely exemplar-based in-context generalization we saw in our synthetic experiment.

\paragraph{Smaller models are less rule-based.}  We investigate this further by evaluating language models of different sizes. We measure `rule-ness' as how much more likely the model is the generalize along the predictive feature (as supported by a sparse rule) in the partial exposure condition, compared to the corresponding (model-specific) control condition. This corresponds to the difference between the purple bar and dotted control in Figs \ref{fig:from_scratch} and \ref{fig:LM}. We find that smaller models (1B and 7B parameters) are steadily less rule-based, with the 1B model effectively performing exact exemplar-based generalization -- similar to those trained from scratch (Fig \ref{fig:LM:sizes}, further details in Appendix \ref{fig:LM_allsizes}).


\subsection{In-context generalization can be made more rule-based with pre-training.}

We did not observe the same effect of scale in transformers that were trained from scratch on synthetic data -- increasing number of layers, number of attention heads, and number of classes did not lead to more rule-basedness. This may be because we were not able to achieve the necessary scale with those experiments or due to synergistic effects between scale and the type of training data.

To evaluate the role of training data, we evaluated in-context generalization on a transformer trained on synthetic stimuli where the query explicitly required rule-based classification (more details in Appendix \ref{app:pretrain_rule_based}). With this training data, the transformer learns rule-based generalization to held-out sequences \ref{fig:from_scratch:train_on_rule_based}. Thus, while transformers exhibit inherent bias towards exemplar-based generalization from context, when trained on data that ecnourages rule-based generalization from context, they can learn to do so. This supports the interpretation that pretrained language model show rule-based generalization because natural language contains implicitly rule-based data, but crucially, this structure only seems to be picked up and used by larger models.

\section{Conclusions}

We found distinct patterns of generalization when transformers generalize from information stored in weights vs in context. When trained on synthetic data, generalization from in-weights information is completely rule-based, whereas generalization from in-context information is almost entirely exemplar-based. However, a pre-trained language model is surprisingly rule-based when generalizing from in-context information, and this is increasingly true for larger models. We find that it is indeed possible to induce rule-based generalization from in-context information by pretraining a transformer on an explicitly rule-based classification problem. Together, these findings support the possibility that natural language data (perhaps because of its combinatorial nature) provides a strong learning pressure towards rule-like generalization, which works in concert with model scale.


\bibliographystyle{plainnat}
\bibliography{refs}

\appendix

\section{Experiment details: Trained-from-scratch transformers}
\label{app:from_scratch_details}

\begin{figure*}[htb]
    \centering
    \begin{subfigure}[t]{0.4\textwidth}
        \centering
        \caption{Transformer inputs and outputs.}
        \includegraphics[width=\textwidth]{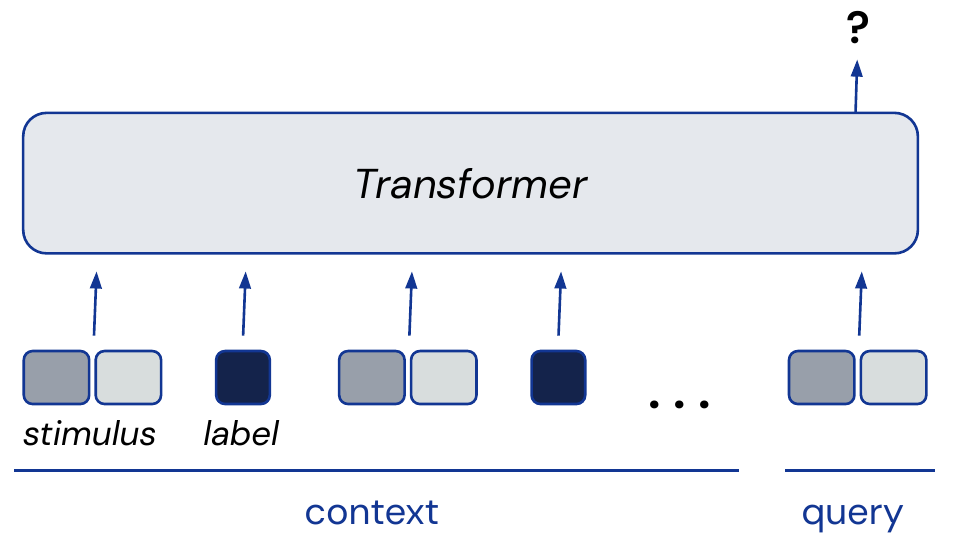}
        \label{app:fig:transformer}
    \end{subfigure}
    \hfill
    \begin{subfigure}[t]{0.58\textwidth}
        \centering
        \caption{Example sequences: Evaluating generalization from context.}
        \includegraphics[width=\textwidth]{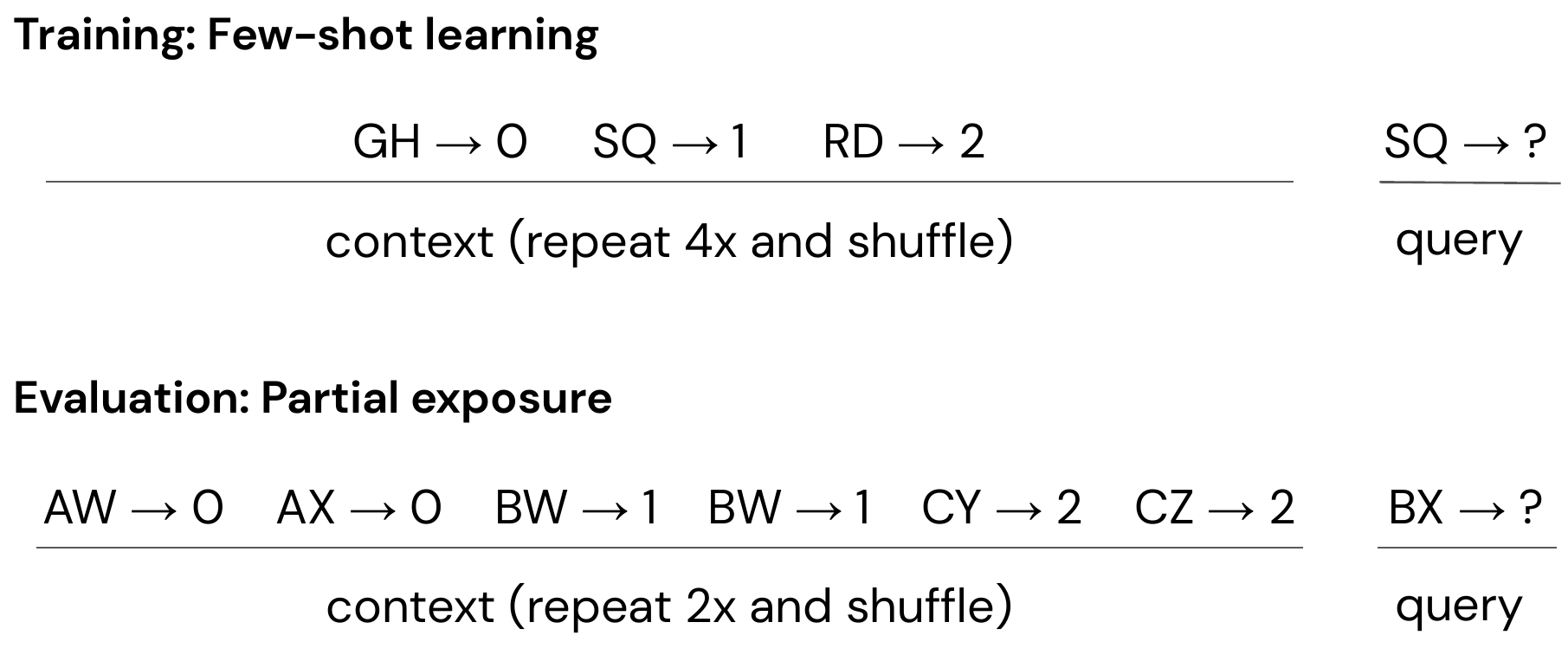}
        \label{app:fig:in_context_seqs}
    \end{subfigure}
    \hfill
    \begin{subfigure}[t]{0.3\textwidth}
        \centering
        \caption{Stimulus examples.}
        \includegraphics[width=\textwidth]{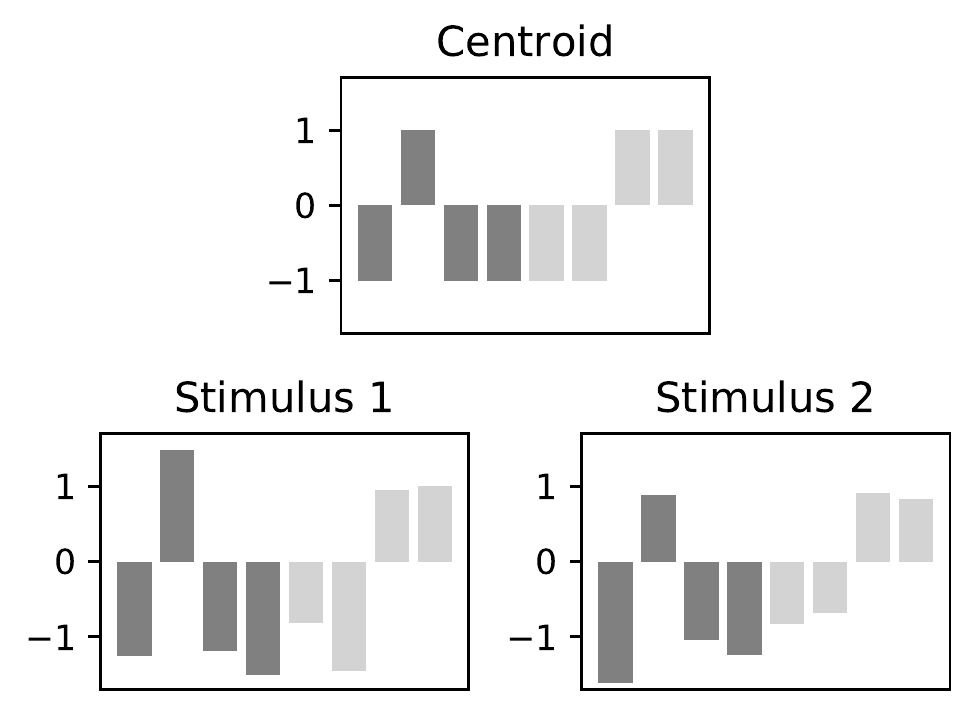}
        \label{app:fig:stimulus_examples}
    \end{subfigure}
    \hfill
    \begin{subfigure}[t]{0.68\textwidth}
        \centering
        \caption{Example sequences: Evaluating generalization from weights.}
        \includegraphics[width=\textwidth]{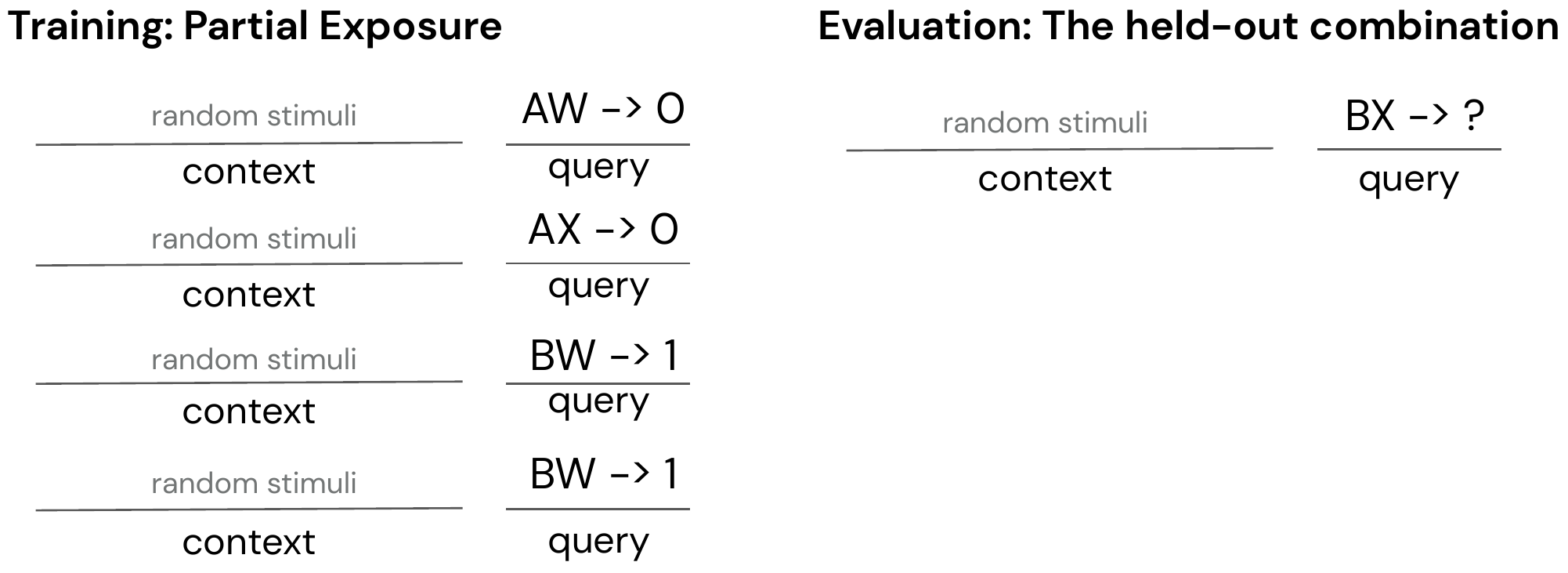}
        \label{app:fig:in_weights_seqs}
    \end{subfigure}
    \hfill
    \vskip-1em
    \caption{
    \textbf{(\subref{app:fig:transformer})} Sequences of alternating stimuli and labels are passed to a transformer. Each sequence consists of a ``context'' (12 stimulus-label pairs) and a ``query'' stimulus. The model is trained to minimize the loss on the query prediction.
    \textbf{(\subref{app:fig:in_context_seqs})} To evaluate generalization from context, the model is first pretrained to perform in-context learning by training on few-shot sequences; stimulus classes and labels are randomly chosen for every sequence, so that the model must perform few-shot learning from context. Inductive biases are evaluated on ``partial exposure'' sequences, where one combination is held out for evaluation (``BX"). Consistent selection of the label associated with ``B" indicates a rule-based bias, while equal selection of the labels associated with ``A" and ``B" indicates an exemplar-based bias (since ``BX" is equally similar to ``AX" and ``BW").
    \textbf{(\subref{app:fig:stimulus_examples})} Each stimulus consists of two subvectors concatenated together into a single token. Each subvector belongs a particular class, and each class is characterized by a different centroid. The subvectors are sampled from a multivariate normal centered on that centroid. The subvectors have length 32, but here we only show 4 values.
    \textbf{(\subref{app:fig:in_weights_seqs})} To evaluate generalization from weights, the model is instead trained directly on partial exposure data, and inductive biases are evaluated on the held-out combination. (The context consisted of random samplings of the stimulus classes, and were irrelevant to the query prediction.)
    }
    \label{app:fig:from_scratch_examples}
    \vskip-0.5em
\end{figure*}

\subsection{Subvector stimuli}
 Each subvector belongs a particular ``subvector class'', and each subvector class is characterized by a different centroid. The subvectors are sampled from a multivariate normal centered on that centroid. A ``stimulus class'' is the concatenation of two subvector classes, and a stimulus is a sampling from a stimulus class.
Example stimuli are shown in Fig \ref{app:fig:from_scratch_examples}. 
This design is a 64-dimensional generalization of the 2-D classification example from \cite{dasgupta_distinguishing_2022}, and 
ensures that there is no a priori bias towards one feature or another. 

Number of features per stimulus: 2  |  
Feature length: 32  |  
 Number of classes per feature: 10  |  
Number of values per class: 100  |  
 Covariance scaling on the per-class normal distributions: 0.1

\subsection{Pretraining for few-shot learning}
\label{app:pretrain_fsl}
To pretrain the model for in-context learning, we pretrained the model on few-shot (4-shot 3-way) sequences, i.e. sequences for which the context consisted of 3 different stimulus classes each repeated 4 times, and where the query class was one of those 3 classes. The classes and labels were randomly assigned for each sequence. 

\subsection{Evaluating inductive biases}

Models were trained and evaluated as shown in Figs \ref{app:fig:in_context_seqs} and \ref{app:fig:in_weights_seqs}. 

An additional label ("2", in the example in Fig \ref{app:fig:in_context_seqs}) and corresponding examples were included in the partial exposure sequences, to ensure that if the model is equally likely to select the labels associated with "A" and "B", it is because of those examples' similarity to the query stimulus, rather than because the model is selecting labels at chance. This was similarly done for training on partial exposure data in weights (not shown in Fig \ref{app:fig:in_weights_seqs}).

Note also that the \texttt{BW} stimulus was shown twice as often as other stimuli in the partial exposure data, as was done in \cite{dasgupta_distinguishing_2022}, in order to ensure that there is no bias induced by having one label more frequent than another. We also ran experiments where \texttt{BW} was not repeated, and the pattern of results was the same.

\subsection{Pretraining for rule-based generalization}
\label{app:pretrain_rule_based}
To pretrain the model for rule-based generalization, we used the partial exposure sequences shown in \ref{app:fig:in_context_seqs}, but for training \textit{as well as} for evaluation. In training, the label for the query \texttt{BX} was always the same as the label associated with \texttt{BW} in context (note that the stimuli and labels are randomly assigned, so that \texttt{BX} could be manifested as any of the stimulus classes, and the query label could be any of [0, 1, 2].)

\subsection{Architecture, training, and evaluation details}
Num layers: 12  |  
Embedding size: 64  |  
Optimizer: Adam  |  
Batch size: 32  |  
Learning rate schedule: Linear warmup and square root decay, described as \texttt{min(3e-4 / 4000 * global\_step, power(4000, 0.5) * 3e-4 * power(global\_step, -0.5))}

For each experiment, we ran 16 TPUv3 cores and 4 v100 GPU cores for 200,000 training steps.

 Models are evaluated on evaluation data throughout training, and bar plots in Fig \ref{fig:from_scratch} show evaluation outputs averaged over the last half of training (100k-200k steps). Error bars indicate 1.96 standard errors across 10 training runs (there is zero variance for the results in Figs \ref{fig:from_scratch:in-weights} and \ref{fig:from_scratch:train_on_rule_based}).

\section{Experiment details: Large language model experiments}
\label{app:LLM_details}

\begin{figure*}[htb]
    \centering
    \includegraphics[width=0.9\textwidth]{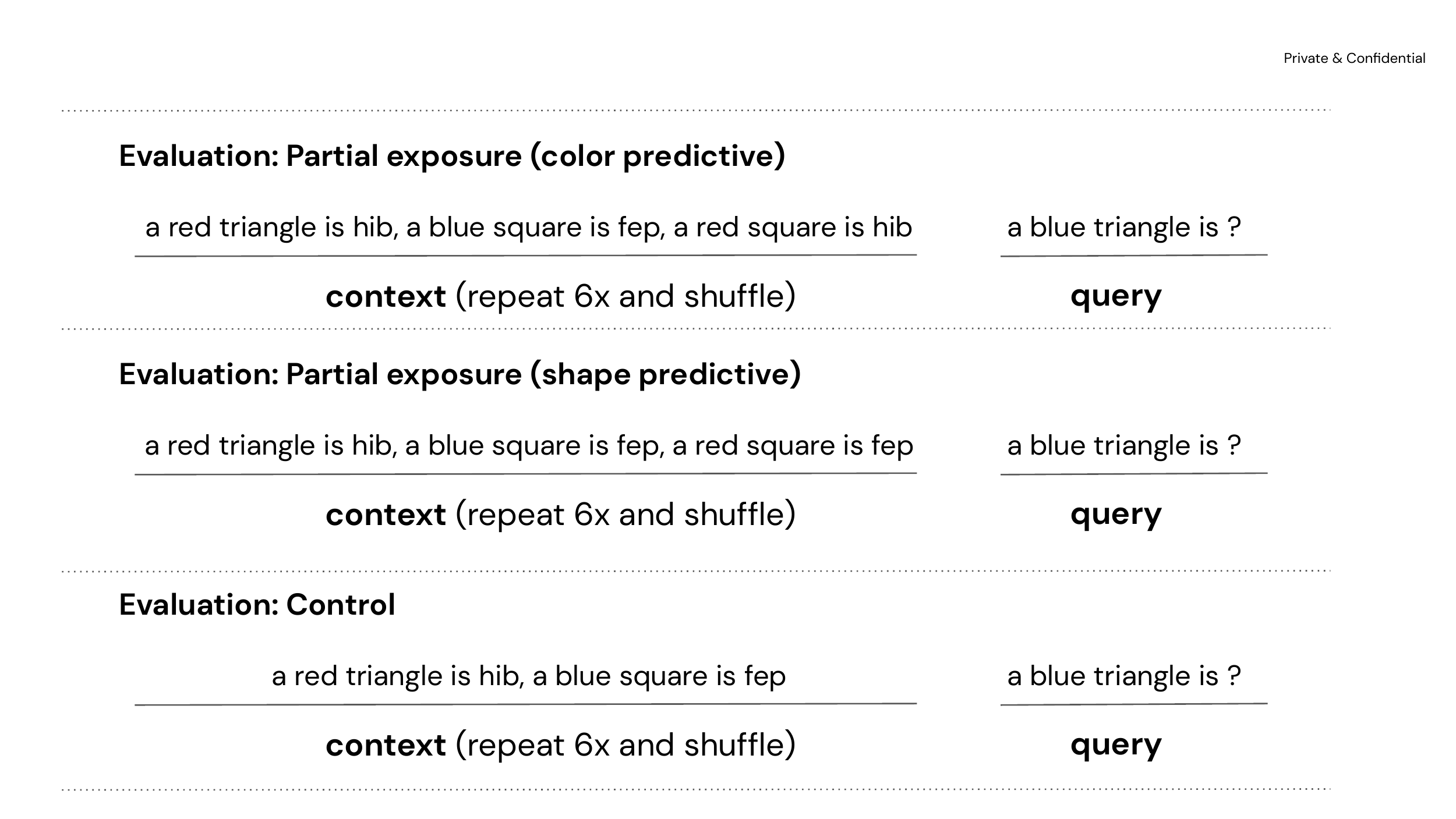}
    \caption{To evaluate generalization from context in a pretrained language model, the model is evaluated on partial exposure sequences where the features are instead text features (shape and color words). The control condition allows us to evaluate the model's baseline bias towards shape or color.}
    \label{app:fig:LM_seqs}
    \vskip-0.5em
\end{figure*}

See Fig \ref{app:fig:LM_seqs} for example evaluation sequences.

In order to account for potential low-level word-order effects, we evaluated on four different formats for the stimuli (`red circle', `a circle that is red', `an object that is circular and red', `an object that is red and circular'). We found no significant differences in qualitative patterns across the different word orderings, so we report the average across all four formats in our results.

\begin{figure*}[tb]
    \centering
    \begin{subfigure}[t]{0.3\textwidth}
        \centering
        \caption{Control condition.}  
        \includegraphics[width=\textwidth]{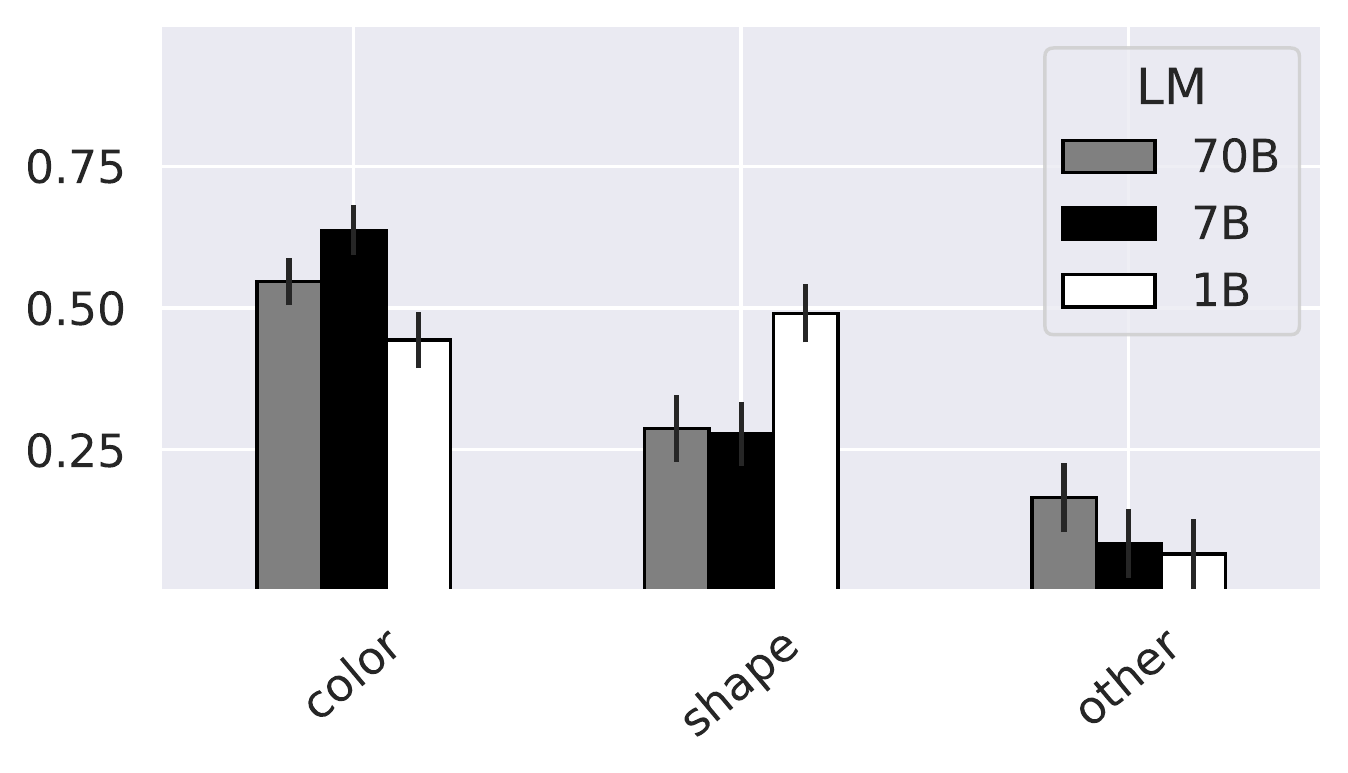}
        \label{fig:LM:control_allsizes}
    \end{subfigure}
    \hfill
    \begin{subfigure}[t]{0.3\textwidth}
        \centering
        \caption{Shape predictive.}  
        \includegraphics[width=\textwidth]{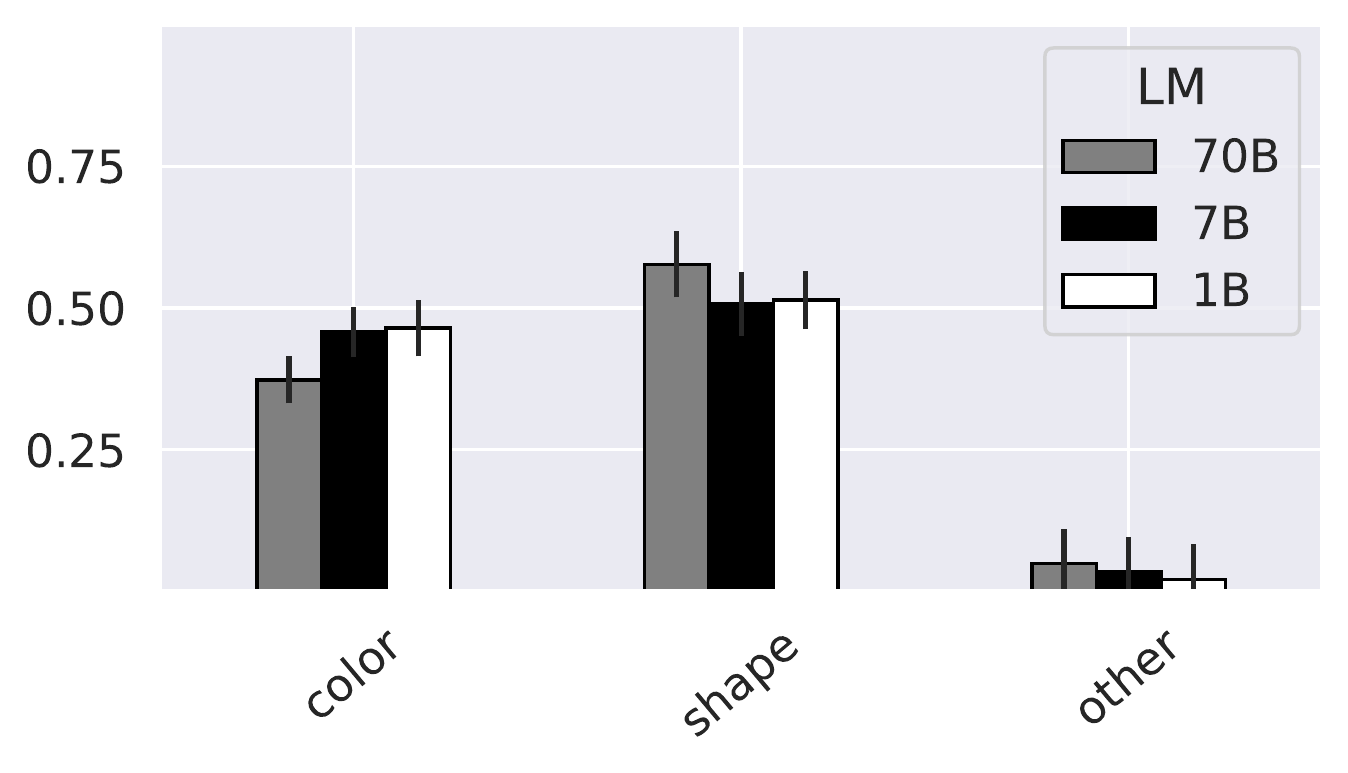}
        \label{fig:LM:shape_allsizes}
    \end{subfigure}
    \hfill
    \begin{subfigure}[t]{0.3\textwidth}
        \centering
        \caption{Color predictive.}  
        \includegraphics[width=\textwidth]{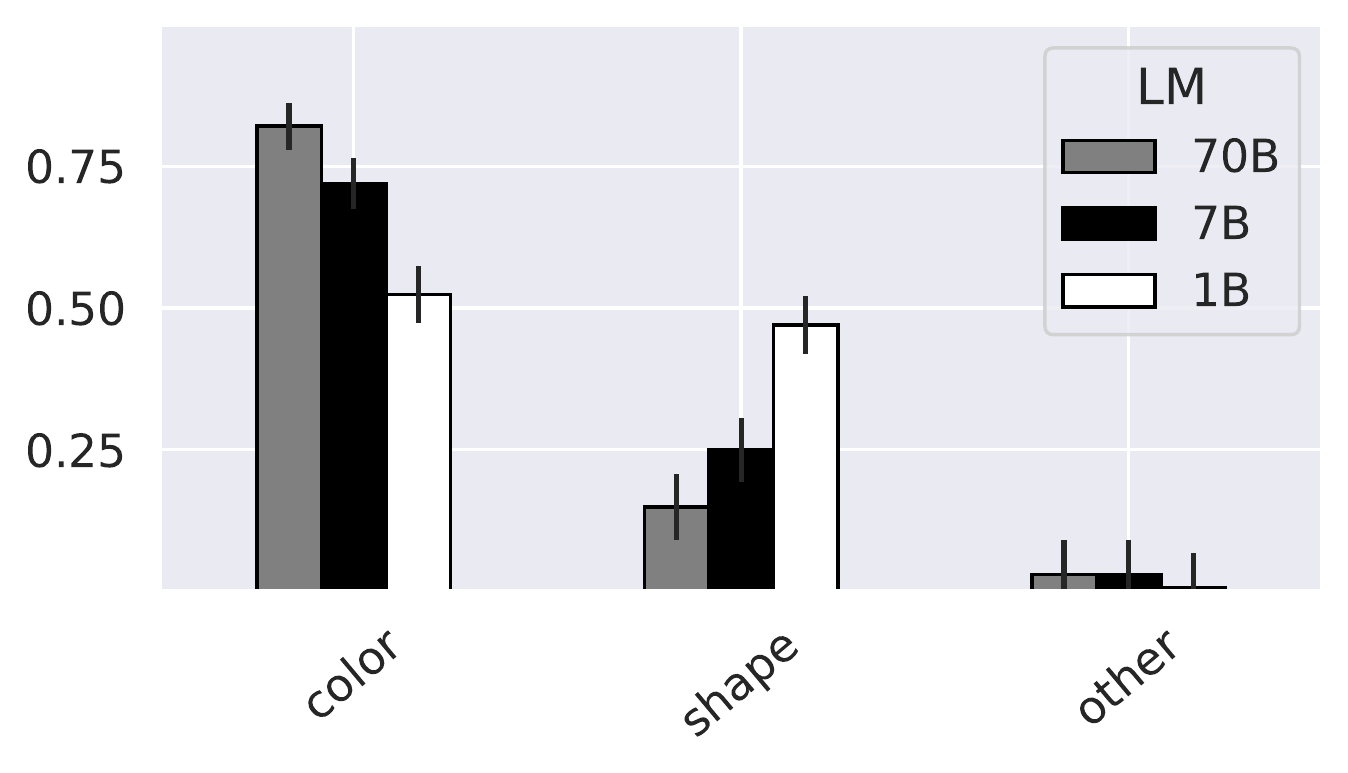}
        \label{fig:LM:color_allsizes}
    \end{subfigure}
    \hfill
    \vskip-1em
    \caption{Results on the pretrained language models for different sizes.}
    \label{fig:LM_allsizes}
    \vskip-0.5em
\end{figure*}

We also include more detailed information about the experiments using different model sizes in Fig \ref{fig:LM_allsizes}. These show the raw performances (including model-specific controls) on all model sizes and feature sets.

\end{document}